\documentclass[10pt]{article}

\usepackage[margin=1in]{geometry}
\usepackage{times}
\usepackage{microtype}
\usepackage{graphicx}
\usepackage{booktabs}
\usepackage{multirow}
\usepackage{amsmath,amssymb,amsfonts}
\usepackage{url}
\usepackage{xcolor}
\usepackage{caption}
\usepackage{subcaption}
\usepackage{enumitem}
\usepackage{array}
\usepackage{makecell}
\usepackage{hyperref}
\hypersetup{
    colorlinks=true,
    linkcolor=blue,
    citecolor=blue,
    urlcolor=blue
}
\graphicspath{{figures/}}

\title{Echo-LoRA: Parameter-Efficient Fine-Tuning via Cross-Layer Representation Injection}

\author{
Yihang Peng$^{1}$ \quad Peng Jin$^{2}$ \quad Jie Gong$^{1}$ \quad Xingyuan Chen$^{2}$\\
Lingjiao Xu$^{2}$ \quad Ning Su$^{2}$ \quad Yan Ran$^{1}$\\[2mm]
$^{1}$School of Computer Science and Software Engineering, Southwest Petroleum University\\
$^{2}$School of Electronic Information and Artificial Intelligence, Leshan Normal University
}

\date{}

\begin{document}
\maketitle

\begin{abstract}
Parameter-efficient fine-tuning (PEFT) has become a practical route for adapting large language models to downstream tasks, with LoRA-style methods being particularly attractive because they are inexpensive to train and easy to deploy. Most LoRA variants, however, revise the update rule within the weight space of each layer and leave the intermediate representations formed by deeper layers largely unused. We propose Echo-LoRA, a cross-layer representation injection method for parameter-efficient fine-tuning. During training, Echo-LoRA collects boundary hidden states from deeper source layers, aggregates them into a sample-level echo representation, and uses lightweight projection and gating networks to inject the resulting signal into shallow LoRA or DoRA modules. Answer-only masking, masked distillation, and stochastic routing are used to keep this auxiliary path stable and to reduce the gap between training and inference. On eight commonsense reasoning benchmarks, Echo-LoRA exceeds the reported LoRA baselines by 5.7 percentage points on average across LLaMA-7B, LLaMA2-7B, and LLaMA3-8B. Under reproduced LoRA baselines in our unified implementation, the average gain is 3.0 points; when combined with DoRA, the gain is 2.7 points. The Echo path is discarded after training, so the deployed model keeps the original low-rank LoRA/DoRA form and adds neither inference-time parameters nor inference computation.
\end{abstract}

\section{Introduction}

Large language models (LLMs) now serve as a common foundation for natural language understanding, text generation, and complex reasoning~\cite{touvron2023llama,touvron2023llama2,dubey2024llama3}. A pretrained model is rarely used unchanged in downstream applications; it usually needs to be adapted to a new task distribution or instruction format~\cite{taori2023alpaca}. Full fine-tuning remains costly for contemporary LLMs, both in training memory and in storage for task-specific checkpoints, which keeps efficient adaptation a practical bottleneck~\cite{liu2024dora,houlsby2019parameter}.

PEFT methods reduce this burden by training only a small set of additional or reparameterized variables. Adapter modules, prompt or prefix tuning, and low-rank updates represent the main families of such methods~\cite{houlsby2019parameter,li2021prefix,lester2021power,liu2022ptuning}. LoRA is often used as a default low-rank baseline: it represents the weight update with two low-rank matrices, trains efficiently, and can be merged into the frozen weights before deployment~\cite{hu2022lora}. Later variants have mostly refined this weight-update view, for instance through adaptive rank allocation, vector-based random-matrix adaptation, or magnitude-direction decomposition~\cite{zhang2023adalora,kopiczko2024vera,liu2024dora}.

This view leaves one question underexplored: during adaptation, what information should a trainable layer receive? Probing studies suggest that Transformer layers do not play identical roles. Shallow layers are more closely tied to lexical, syntactic, and local patterns, while deeper layers tend to encode more abstract semantic and task-relevant information~\cite{alain2017understanding,tenney2019bert}. A shallow LoRA module, by design, updates its layer from the representations available at that point in the forward pass; it has no direct access to semantic states that appear later in the network. We argue that this separation can be limiting for tasks that depend on global judgment, commonsense integration, or structured generation.

Echo-LoRA is built around this observation. During training, we extract answer-boundary representations from deeper source layers, aggregate them into a sample-level echo representation, and inject that representation into shallow LoRA/DoRA modules using small projection and gating networks. The design gives shallow adaptation modules access to a compact signal derived from deeper semantic states. Since such a cross-layer path can also introduce spurious dependencies, we use answer-only masking, masked distillation, and stochastic routing to keep the auxiliary signal controlled.

Our contributions are as follows. We introduce Echo-LoRA, a training-time cross-layer injection mechanism that feeds answer-boundary representations from deeper layers into shallow LoRA/DoRA adaptation modules. We pair this mechanism with answer-only masking, masked distillation, and stochastic routing, so that the auxiliary path helps optimization without being required at inference time. We evaluate Echo-LoRA on LLaMA-7B, LLaMA2-7B, and LLaMA3-8B. On eight commonsense reasoning datasets, it improves the average score by 5.7 points over reported LoRA baselines and by 3.0 points over reproduced LoRA baselines in our unified implementation. Echo-DoRA improves the corresponding DoRA baseline by 2.7 points. Additional experiments on mathematical reasoning, code generation, and multitask knowledge understanding show similar positive trends.

\section{Related Work}

\subsection{Parameter-Efficient Fine-Tuning for LLMs}

Scaling pretrained language models makes full fine-tuning increasingly expensive and harder to maintain in deployment. PEFT methods respond by keeping the backbone mostly frozen and updating a small subset of trainable parameters~\cite{houlsby2019parameter}. Adapter-based methods insert compact trainable modules into the original network~\cite{houlsby2019parameter,mahabadi2021compacter}, whereas prompt- and prefix-tuning optimize continuous vectors that steer generation~\cite{li2021prefix,lester2021power,liu2022ptuning}.

Low-rank updates form another widely used PEFT family. LoRA approximates the full weight update by two low-rank matrices and obtains strong performance at modest training cost~\cite{hu2022lora}. AdaLoRA adjusts rank budgets according to parameter importance~\cite{zhang2023adalora}; VeRA fixes random matrices and trains scaling vectors~\cite{kopiczko2024vera}; DoRA decomposes pretrained weights into magnitude and direction components to better mimic full fine-tuning~\cite{liu2024dora}. These methods mainly refine how local weights are updated. Echo-LoRA instead asks whether hidden states from deeper layers can serve as useful conditioning signals for shallow adaptation modules.

\subsection{Intermediate Representations and Cross-Layer Information}

Transformer hidden states vary substantially with depth. Probing and interpretability studies have found that shallow layers tend to encode local lexical and syntactic patterns, while deeper layers capture more abstract semantic information~\cite{alain2017understanding,tenney2019bert}. This layered structure suggests that depth itself can provide a useful source of training signal.

Layer differences have also been used at inference time. DoLa, for example, contrasts output distributions from different layers during decoding to improve factuality and accuracy~\cite{chuang2024dola}. These results motivate treating intermediate states as reusable signals rather than disposable computation traces. Echo-LoRA follows this line of thought in the fine-tuning stage, where we use deeper representations as a training-time auxiliary condition for shallow PEFT modules.

\subsection{Training Stability Strategies}

Randomized computation paths and selective supervision are standard tools for improving robustness. Stochastic Depth, for instance, drops layers during training to reduce dependence on a fixed path~\cite{huang2016stochastic}. Instruction tuning typically ignores prompt tokens in the loss and backpropagates only through target-answer positions~\cite{taori2023alpaca}. Echo-LoRA uses a related principle: because the Echo branch adds a second, training-only path, we restrict where the injected signal appears and randomize when the path is active.

\section{Method}

\subsection{Problem Definition and LoRA Preliminaries}

Let the input sequence be $X=(x_1,\ldots,x_T)$. In instruction tuning, the sequence is usually the concatenation of a prompt and an answer. We denote the answer-token positions by $\mathcal{A}$. The model predicts these answer tokens autoregressively conditioned on the prompt, while prompt positions are typically excluded from the loss.

Consider a target linear transformation in a Transformer block with input $\mathbf{u}\in\mathbb{R}^{d_{\text{in}}}$ and output $\mathbf{o}\in\mathbb{R}^{d_{\text{out}}}$. LoRA keeps the pretrained weight $\mathbf{W}$ frozen and learns a low-rank update:
\begin{equation}
\mathbf{o}=\mathbf{W}\mathbf{u}+\Delta\mathbf{W}\mathbf{u},\quad
\Delta\mathbf{W}=\frac{\alpha}{r}\mathbf{B}\mathbf{A},
\end{equation}
where $\mathbf{W}\in\mathbb{R}^{d_{\text{out}}\times d_{\text{in}}}$ is the frozen pretrained weight, $\mathbf{A}\in\mathbb{R}^{r\times d_{\text{in}}}$ and $\mathbf{B}\in\mathbb{R}^{d_{\text{out}}\times r}$ are trainable low-rank matrices, $r$ is the rank, and $\alpha$ is a scaling coefficient.

This update remains local to the target layer. It does not condition shallow trainable modules on representations that appear later in the network. If deeper hidden states contain task-relevant semantic information, using them as a training-time signal may improve the adaptation of shallow modules.

\subsection{Overall Framework}

Echo-LoRA uses deeper representations as auxiliary conditions for shallow PEFT modules. At routed training steps, we collect hidden states at answer-boundary positions from deeper source layers, aggregate them into a sample-level echo representation, and inject the resulting signal into shallow target LoRA/DoRA modules through projection and gating networks.

Let $\mathcal{S}$ be the source-layer set and $\mathcal{T}$ the target-layer set, with source layers placed deeper than target layers. For sample $b$, $t_b^{\star}$ denotes the boundary position immediately before the answer region. We extract source-layer hidden states at this position and average them:
\begin{equation}
\mathbf{z}_b=\frac{1}{|\mathcal{S}|}\sum_{l\in\mathcal{S}}\mathbf{h}^{(l)}_{b,t_b^{\star}}.
\end{equation}
Here $\mathbf{h}^{(l)}_{b,t_b^{\star}}$ is the hidden state of layer $l$ at position $t_b^{\star}$. We use this boundary position because it is expected to summarize the prompt context before answer generation begins. In implementation, the first forward pass produces the source representation, and the second pass uses it as a stop-gradient condition; the injection branch does not backpropagate through the source hidden states obtained in the first pass.

Given $\mathbf{z}_b$, Echo-LoRA first normalizes it and then computes an injection vector through projection and gating networks. For a target layer $l\in\mathcal{T}$ and target module $m$, the computation is:
\begin{equation}
\bar{\mathbf{z}}_b=\mathrm{Norm}(\mathbf{z}_b),
\end{equation}
\begin{equation}
\mathbf{e}_b^{(l,m)}=\mathbf{W}^{(l,m)}_2\tanh\left(\mathbf{W}^{(l,m)}_1\bar{\mathbf{z}}_b\right),
\end{equation}
\begin{equation}
\mathbf{g}_b^{(l,m)}=\sigma\left(\mathbf{U}^{(l,m)}_2\tanh\left(\mathbf{U}^{(l,m)}_1\bar{\mathbf{z}}_b\right)+\mathbf{b}^{(l,m)}\right),
\end{equation}
\begin{equation}
\boldsymbol{\delta}_b^{(l,m)}=\lambda^{(l,m)}\left(\mathbf{e}_b^{(l,m)}\odot\mathbf{g}_b^{(l,m)}\right).
\end{equation}
The projection parameters $\mathbf{W}^{(l,m)}_1$ and $\mathbf{W}^{(l,m)}_2$ map the deep representation into the target-module output space through a small bottleneck. The gating parameters $\mathbf{U}^{(l,m)}_1$, $\mathbf{U}^{(l,m)}_2$, and $\mathbf{b}^{(l,m)}$ filter the injected signal in a sample- and module-dependent manner. The scalar $\lambda^{(l,m)}$ is a learnable scale, $\odot$ denotes element-wise multiplication, and $\sigma(\cdot)$ is the sigmoid function. We initialize the gate bias negatively, so the Echo branch begins with a weak activation and is less likely to dominate early updates.

The final injected target-module output is
\begin{equation}
\widetilde{\mathbf{o}}_{b,t}^{(l,m)}=\mathbf{o}_{b,t}^{(l,m)}+r_kM_{b,t}\boldsymbol{\delta}_b^{(l,m)},
\end{equation}
where $\mathbf{o}_{b,t}^{(l,m)}$ is the original module output, $\widetilde{\mathbf{o}}_{b,t}^{(l,m)}$ is the injected output, $M_{b,t}\in\{0,1\}$ is the answer-region mask, and $r_k\in\{0,1\}$ is the stochastic routing variable at step $k$. Since the source layers come after the target layers in the forward computation, training uses two passes: an Echo-off pass that extracts the source boundary representation, followed by an Echo-on pass that injects this representation into shallow target modules and computes the losses.

\begin{figure}[t]
\centering
\includegraphics[width=0.95\textwidth]{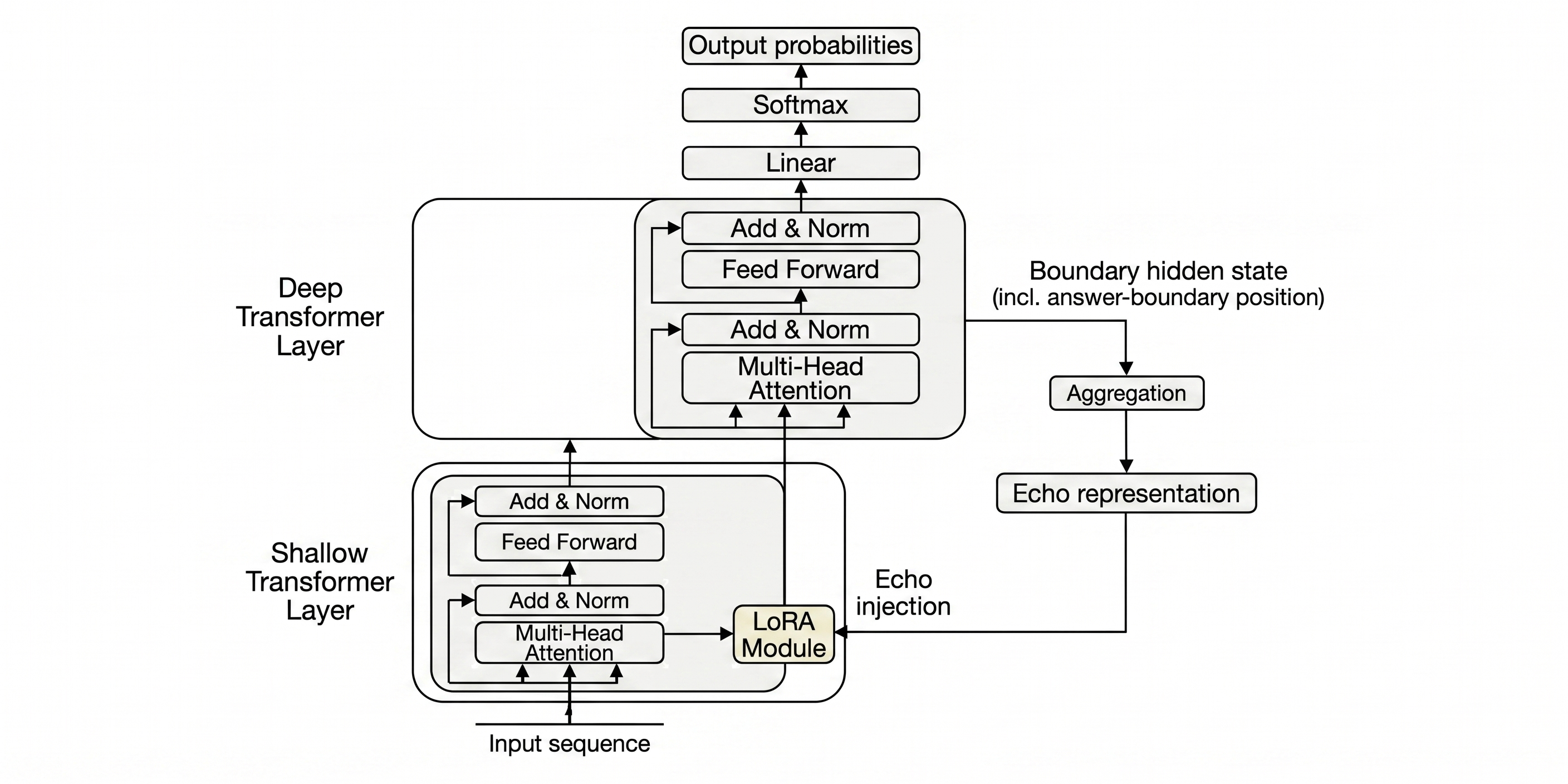}
\caption{Overall framework of Echo-LoRA. Boundary hidden states are extracted from deeper layers, aggregated into an echo representation, and injected into shallow LoRA modules during training.}
\label{fig:framework}
\end{figure}

\subsection{Answer-Only Selective Injection}

Applying the echo signal to every token would also perturb prompt positions, which normally do not contribute to the language-modeling loss. We restrict Echo injection to supervised answer positions for this reason.

For the $b$-th sample, the binary mask is constructed from the supervised language-modeling labels:
\begin{equation}
M_{b,t}=\begin{cases}
1, & y_{b,t}\neq -100,\\
0, & \text{otherwise}.
\end{cases}
\end{equation}
Here $y_{b,t}$ is the supervised label at position $t$. Positions ignored by the language-modeling loss are typically marked as $-100$, so the mask retains Echo injection only where answer tokens are predicted.

\subsection{Stochastic Routing}

Keeping the Echo path active throughout training can make the model rely too heavily on a branch that will be absent at inference time. Echo-LoRA uses stochastic routing to address this mismatch.

At training step $k$, a Bernoulli variable decides whether the Echo path is enabled:
\begin{equation}
r_k\sim \mathrm{Bernoulli}(p_k).
\end{equation}
All target modules share the same routing state within a training step. The routing probability follows a linear decay:
\begin{equation}
p_k=p_{\text{start}}+\frac{k}{K-1}\left(p_{\text{end}}-p_{\text{start}}\right),
\end{equation}
where $K$ is the total number of training steps and $p_{\text{start}}>p_{\text{end}}$. Early training exposes the target layers more often to the source-layer signal, while later training gradually shifts the model back toward the deployable Echo-off path.

\subsection{Training Objective and Deployment}

Training starts from an Echo-off forward pass, which gives the supervised loss $\mathcal{L}_{\text{off}}$. When routing activates the Echo branch, a second Echo-on pass produces $\mathcal{L}_{\text{on}}$ and a masked distillation loss $\mathcal{L}_{\text{kd}}$ on answer positions. The objective is
\begin{equation}
\mathcal{L}_t=\mathcal{L}_{\text{off}}+r_t\left(\mathcal{L}_{\text{on}}+\lambda_{\text{kd}}\mathcal{L}_{\text{kd}}\right),
\end{equation}
where $r_t$ is the routing variable and $\lambda_{\text{kd}}$ is the distillation weight. If $r_t=0$, only the base path is optimized; if $r_t=1$, both paths are optimized and their answer-region predictions are constrained to remain close.

Let
$\mathbf{q}_{b,t}^{\mathrm{on}}=\operatorname{softmax}(\mathbf{p}_{b,t}^{\mathrm{on}}/\tau)$ and
$\mathbf{q}_{b,t}^{\mathrm{off}}=\operatorname{softmax}(\mathbf{p}_{b,t}^{\mathrm{off}}/\tau)$. The distillation loss is
\begin{equation}
\mathcal{L}_{\mathrm{kd}}
=\frac{\tau^2}{|\mathcal{A}|}
\sum_{(b,t)\in\mathcal{A}}
\operatorname{KL}\left(
\mathbf{q}_{b,t}^{\mathrm{on}}\,\middle\|\,\mathbf{q}_{b,t}^{\mathrm{off}}
\right).
\end{equation}
The Echo-on branch serves as the teacher and is detached in the distillation term. Following standard knowledge distillation, the factor $\tau^2$ keeps the gradient scale comparable across temperatures. Unless otherwise stated, we set $\lambda_{\text{kd}}=1.0$ and $\tau=2.0$.

Echo modules are training-only components. At deployment, we disable the Echo path, so no echo extraction, projection, or injection is executed. The deployable model has the same inference structure as standard LoRA/DoRA; auxiliary Echo parameters may remain in checkpoints for analysis, but they are not used during generation.

\section{Experiments}

\subsection{Experimental Setup}

We use LLaMA-7B, LLaMA2-7B, and LLaMA3-8B as the main backbones, with LLaMA2-13B included in extended experiments. The commonsense reasoning setup follows the benchmark suite used by DoRA~\cite{liu2024dora}. To make the source of each comparison explicit, we separate main-table comparisons against published baselines from reproduced LoRA comparisons under our unified implementation, which are reported in Appendix~\ref{app:lora-full}.

Table~\ref{tab:datasets} summarizes the datasets. The first eight datasets form the commonsense reasoning suite: their training sets are mixed during training, and their evaluation sets are tested separately. GSM8K, HumanEval, and MMLU are used for extended evaluation on mathematical reasoning, code generation, and multitask knowledge understanding.

\begin{table}[t]
\centering
\caption{Datasets used in the experiments.}
\label{tab:datasets}
\small
\begin{tabular}{llrr}
\toprule
Dataset & Task Type & Train/Dev Size & Test Size \\
\midrule
BoolQ & Yes/no reasoning & 9,427 & 3,270 \\
PIQA & Physical commonsense & 16,113 & 1,838 \\
SIQA & Social commonsense & 33,410 & 1,954 \\
HellaSwag & Situation continuation & 39,905 & 10,042 \\
WinoGrande & Coreference reasoning & 40,398 & 1,267 \\
ARC-e & Science QA (Easy) & 2,251 & 2,376 \\
ARC-c & Science QA (Challenge) & 1,119 & 1,172 \\
OBQA & Open-book commonsense QA & 4,957 & 500 \\
GSM8K & Mathematical reasoning & 7,473 & 1,319 \\
HumanEval & Code generation & -- & 164 \\
MMLU & Multitask knowledge & 285 & 14,042 \\
\bottomrule
\end{tabular}
\end{table}

Unless otherwise stated, our experiments use rank $r=16$, scaling coefficient $\alpha=32$, and apply LoRA/DoRA to the attention projections \texttt{q\_proj}, \texttt{k\_proj}, \texttt{v\_proj}, and \texttt{o\_proj}. The learning rate is $2\times10^{-4}$, LoRA dropout is 0.05, training lasts for 3 epochs, the maximum sequence length is 256, and bf16 training is used. For four-GPU training, each GPU uses batch size 2, resulting in total batch size 16.

For the Echo mechanism, the default source layers are $[-8,-7,-6,-5]$, the default target layers are $[4,5,6,7]$, and injection is applied to \texttt{q\_proj} and \texttt{v\_proj}. The bottleneck dimension is 64. Stochastic routing uses $p_{\text{start}}=1.0$ and $p_{\text{end}}=0.2$. Answer-only masking and masked distillation are enabled by default, with $\lambda_{\text{kd}}=1.0$ and $\tau=2.0$.

\subsection{Main Commonsense Reasoning Results}

Table~\ref{tab:main-results} reports the LoRA-line results on eight commonsense reasoning datasets. Prefix, Series, Parallel, and LoRA values are taken from the DoRA paper~\cite{liu2024dora}; Echo-LoRA values are obtained by our method. All average scores are computed from the eight displayed task scores.

\begin{table*}[t]
\centering
\caption{Accuracy comparison on eight commonsense reasoning datasets (\%). Prefix, Series, Parallel, and LoRA are reported by DoRA~\cite{liu2024dora}; Echo-LoRA is our method.}
\label{tab:main-results}
\resizebox{\textwidth}{!}{%
\begin{tabular}{llccccccccc}
\toprule
Model & PEFT Method & BoolQ & PIQA & SIQA & HellaSwag & WinoGrande & ARC-e & ARC-c & OBQA & Avg. \\
\midrule
\multirow{5}{*}{LLaMA-7B}
& Prefix & 64.3 & 76.8 & 73.9 & 42.1 & 72.1 & 72.9 & 54.0 & 60.6 & 64.6 \\
& Series & 63.0 & 79.2 & 76.3 & 67.9 & 75.7 & 74.5 & 57.1 & 72.4 & 70.8 \\
& Parallel & 67.9 & 76.4 & 78.8 & 69.8 & 78.9 & 73.7 & 57.3 & 75.2 & 72.2 \\
& LoRA & 68.9 & 80.7 & 77.4 & 78.1 & 78.8 & 77.8 & 61.3 & 74.8 & 74.7 \\
& Echo-LoRA & 63.6 & 82.9 & 77.6 & 93.7 & 84.5 & 85.0 & 70.1 & 81.2 & \textbf{79.8} \\
\midrule
\multirow{2}{*}{LLaMA2-7B}
& LoRA & 69.8 & 79.9 & 79.5 & 83.6 & 82.6 & 79.8 & 64.7 & 81.0 & 77.6 \\
& Echo-LoRA & 72.7 & 83.7 & 80.5 & 94.0 & 85.9 & 87.6 & 74.7 & 84.6 & \textbf{83.0} \\
\midrule
\multirow{2}{*}{LLaMA3-8B}
& LoRA & 70.8 & 85.2 & 79.9 & 91.7 & 84.3 & 84.2 & 71.2 & 79.0 & 80.8 \\
& Echo-LoRA & 75.6 & 90.2 & 82.5 & 96.6 & 89.8 & 93.6 & 82.5 & 89.2 & \textbf{87.5} \\
\bottomrule
\end{tabular}}
\end{table*}

Echo-LoRA improves the reported LoRA average on all three backbones. The gains are 5.1, 5.4, and 6.7 points for LLaMA-7B, LLaMA2-7B, and LLaMA3-8B, respectively, giving an average gain of 5.7 points. The largest gain is observed on LLaMA3-8B, suggesting that the Echo signal is not merely compensating for a weak backbone but can also complement a stronger one.

\subsection{Combining Echo with DoRA}

We next apply the same Echo mechanism to DoRA, yielding Echo-DoRA. Table~\ref{tab:dora-results} compares DoRA and Echo-DoRA on the same eight commonsense reasoning tasks. DoRA values are from the DoRA paper~\cite{liu2024dora}, while Echo-DoRA values are our results. For numerical consistency, the average scores are recomputed from the displayed task scores.

\begin{table*}[t]
\centering
\caption{Accuracy comparison on eight commonsense reasoning datasets (\%). DoRA values are from DoRA~\cite{liu2024dora}; Echo-DoRA is our method.}
\label{tab:dora-results}
\resizebox{\textwidth}{!}{%
\begin{tabular}{llccccccccc}
\toprule
Model & PEFT Method & BoolQ & PIQA & SIQA & HellaSwag & WinoGrande & ARC-e & ARC-c & OBQA & Avg. \\
\midrule
\multirow{2}{*}{LLaMA-7B}
& DoRA & 69.7 & 83.4 & 78.6 & 87.2 & 81.0 & 81.9 & 66.2 & 79.2 & 78.4 \\
& Echo-DoRA & 69.6 & 82.6 & 81.2 & 93.4 & 83.6 & 86.1 & 70.7 & 83.2 & \textbf{81.3} \\
\midrule
\multirow{2}{*}{LLaMA2-7B}
& DoRA & 71.8 & 83.7 & 76.0 & 89.1 & 82.6 & 83.7 & 68.2 & 82.4 & 79.7 \\
& Echo-DoRA & 73.5 & 84.7 & 81.6 & 94.4 & 85.6 & 88.1 & 74.2 & 87.4 & \textbf{83.7} \\
\midrule
\multirow{2}{*}{LLaMA3-8B}
& DoRA & 74.6 & 89.3 & 79.9 & 95.5 & 85.6 & 90.5 & 80.4 & 85.8 & 85.2 \\
& Echo-DoRA & 75.1 & 89.0 & 81.9 & 97.1 & 88.5 & 91.3 & 81.6 & 86.8 & \textbf{86.4} \\
\bottomrule
\end{tabular}}
\end{table*}

Echo-DoRA improves the DoRA average by 2.9, 4.0, and 1.2 points on the three backbones, respectively, for an average gain of 2.7 points. These gains are smaller than those on the LoRA line, which is consistent with the view that Echo provides a complementary signal when the underlying adaptation method is already stronger.

\subsection{Ablation Study}

Ablations are conducted on LLaMA3-8B with the same training configuration. Table~\ref{tab:ablation} reports the average score over the eight commonsense reasoning tasks. The baseline in this table is our reproduced LLaMA3-8B LoRA baseline, which differs from the published LoRA row in Table~\ref{tab:main-results}. Appendix~\ref{app:ablation-full} provides the task-level results.

\begin{table}[t]
\centering
\caption{Ablation results on LLaMA3-8B (average accuracy over eight commonsense reasoning datasets).}
\label{tab:ablation}
\small
\begin{tabular}{lc}
\toprule
Setting & Avg. \\
\midrule
A-0 Reproduced LoRA baseline & 84.7 \\
A-1 w/o Stochastic Routing & 78.7 \\
A-2 Deep $\rightarrow$ Deep & 84.8 \\
A-3 Shallow $\rightarrow$ Shallow & 86.2 \\
A-4 w/o Answer-Only Masking & 87.2 \\
A-5 \texttt{v\_proj} only & 87.2 \\
A-6 w/o Answer-Only Masking + all attention projections & 87.3 \\
A-7 \texttt{q\_proj} only & 87.4 \\
A-8 All attention projections & 87.5 \\
A-9 Full Echo-LoRA & \textbf{87.5} \\
\bottomrule
\end{tabular}
\end{table}

The score drops most sharply when stochastic routing is removed, so we regard routing as the most important stabilizing component in this set of ablations. Deep-to-deep and shallow-to-shallow variants underperform the default deep-to-shallow configuration, supporting our hypothesis that deeper semantic states are most useful when they guide shallower adaptation modules. Removing answer-only masking or using a single injected projection remains competitive but is slightly weaker than the default. Injecting all attention projections matches the default average score, yet it touches more modules and increases training cost; we keep the simpler \texttt{q\_proj}/\texttt{v\_proj} configuration.

\subsection{Extended Task Results}

We also test whether the same mechanism transfers beyond commonsense reasoning. GSM8K and MMLU are evaluated with accuracy, and HumanEval with pass@1. The baseline values in Table~\ref{tab:extended} are from Flat-LoRA~\cite{li2025flatlora}; Echo-LoRA values are our results.

\begin{table}[t]
\centering
\caption{Extended task results (\%). Baselines are from Flat-LoRA~\cite{li2025flatlora}; Echo-LoRA is our method.}
\label{tab:extended}
\small
\begin{tabular}{llcc}
\toprule
Model & Dataset & Baseline & Echo-LoRA \\
\midrule
\multirow{2}{*}{LLaMA2-7B}
& GSM8K & 56.25 & \textbf{58.61} \\
& HumanEval & 24.56 & \textbf{25.78} \\
\midrule
\multirow{2}{*}{LLaMA2-13B}
& MMLU & 52.27 & \textbf{53.90} \\
& HumanEval & 13.78 & \textbf{15.85} \\
\bottomrule
\end{tabular}
\end{table}

Echo-LoRA improves over the corresponding baselines in all four extended settings. The gains are modest but consistent, suggesting that Echo is not tied to a single benchmark family and may be useful when tasks require reasoning, code synthesis, or broad knowledge integration.

\subsection{Discussion}

The Echo gains are larger on LoRA than on DoRA. We hypothesize that standard LoRA leaves more room for auxiliary semantic guidance, while DoRA has already improved the adaptation dynamics through weight decomposition. From the task side, commonsense reasoning, mathematical reasoning, and code generation all require contextual integration and conditional constraints. Our observations suggest that Echo-LoRA helps by giving shallow adaptation modules access to semantic information that is less local than their native layer states.

At deployment, we disable the Echo path. Inference uses the same low-rank update form as LoRA/DoRA and requires no Echo extraction, projection, or injection. The cost is paid during training: whenever routing activates Echo, a second forward pass is needed. With the routing probability decaying from $p_{\text{start}}=1.0$ to $p_{\text{end}}=0.2$, Echo-LoRA uses more training computation than standard LoRA while preserving the inference-time efficiency that makes low-rank adaptation attractive.

\section{Conclusion}

We introduced Echo-LoRA, a parameter-efficient fine-tuning method that uses cross-layer representation injection during training. The method extracts answer-boundary hidden states from deeper source layers, aggregates them into a sample-level echo representation, and injects this representation into shallow LoRA/DoRA modules. Answer-only masking, masked distillation, and stochastic routing are used to keep the auxiliary path stable and compatible with Echo-off inference.

Across three LLaMA backbones and eight commonsense reasoning benchmarks, Echo-LoRA improves the reported LoRA baselines by 5.7 points on average; under our reproduced LoRA implementation, the average gain is 3.0 points. Echo-DoRA improves DoRA by 2.7 points on average, and extended evaluations on GSM8K, HumanEval, and MMLU show additional positive gains. Because the Echo path is removed at inference time, the deployed model retains the standard LoRA/DoRA form and adds no inference parameters or computation.

Taken together, the results indicate that deeper representations can be useful not only for inference-time decoding strategies but also as training-time auxiliary signals for shallow PEFT modules. We view cross-layer information use as a promising direction for future PEFT research, especially when the auxiliary path can be removed before deployment.

\appendix

\section{Additional Results and Discussion}

\subsection{Full Echo-LoRA Results on Commonsense Reasoning}
\label{app:lora-full}

Table~\ref{tab:lora-task-full} gives the task-level LoRA-line results. Alongside the published LoRA values, we report reproduced LoRA baselines from our unified implementation. The gain $\Delta$ is computed as Echo-LoRA minus the reproduced LoRA baseline. Under this protocol, Echo-LoRA improves the average score by 6.2, 0.8, and 1.9 points on LLaMA-7B, LLaMA2-7B, and LLaMA3-8B, respectively, giving an average gain of 3.0 points.

\begin{table*}[t]
\centering
\caption{Full LoRA-line results on commonsense reasoning tasks (\%). $\Delta$ denotes the absolute gain of Echo-LoRA over the reproduced LoRA baseline under our unified implementation.}
\label{tab:lora-task-full}
\resizebox{\textwidth}{!}{%
\begin{tabular}{llccccccccc}
\toprule
Model & PEFT Method & BoolQ & PIQA & SIQA & HellaSwag & WinoGrande & ARC-e & ARC-c & OBQA & Avg. \\
\midrule
\multirow{4}{*}{LLaMA-7B}
& LoRA (reported) & 68.9 & 80.7 & 77.4 & 78.1 & 78.8 & 77.8 & 61.3 & 74.8 & 74.7 \\
& LoRA (reproduced) & 57.1 & 80.5 & 78.6 & 64.4 & 81.8 & 82.9 & 66.1 & 77.4 & 73.6 \\
& Echo-LoRA & 63.6 & 82.9 & 77.6 & 93.7 & 84.5 & 85.0 & 70.1 & 81.2 & 79.8 \\
& $\Delta$ & +6.5 & +2.4 & -1.0 & +29.3 & +2.7 & +2.1 & +4.0 & +3.8 & +6.2 \\
\midrule
\multirow{4}{*}{LLaMA2-7B}
& LoRA (reported) & 69.8 & 79.9 & 79.5 & 83.6 & 82.6 & 79.8 & 64.7 & 81.0 & 77.6 \\
& LoRA (reproduced) & 71.3 & 84.4 & 80.5 & 93.7 & 85.3 & 87.5 & 73.1 & 82.0 & 82.2 \\
& Echo-LoRA & 72.7 & 83.7 & 80.5 & 94.0 & 85.9 & 87.6 & 74.7 & 84.6 & 83.0 \\
& $\Delta$ & +1.4 & -0.7 & +0.0 & +0.3 & +0.6 & +0.1 & +1.6 & +2.6 & +0.8 \\
\midrule
\multirow{4}{*}{LLaMA3-8B}
& LoRA (reported) & 70.8 & 85.2 & 79.9 & 91.7 & 84.3 & 84.2 & 71.2 & 79.0 & 80.8 \\
& LoRA (reproduced) & 67.6 & 90.3 & 82.1 & 96.5 & 88.0 & 92.3 & 82.7 & 85.4 & 85.6 \\
& Echo-LoRA & 75.6 & 90.2 & 82.5 & 96.6 & 89.8 & 93.6 & 82.5 & 89.2 & 87.5 \\
& $\Delta$ & +8.0 & -0.1 & +0.4 & +0.1 & +1.8 & +1.3 & -0.2 & +3.8 & +1.9 \\
\bottomrule
\end{tabular}}
\end{table*}

\subsection{Full Ablation Results on LLaMA3-8B}
\label{app:ablation-full}

Table~\ref{tab:ablation-full} reports the full task-level ablation results on LLaMA3-8B. Average scores are computed from the eight displayed task scores. The results point to stochastic routing as the dominant stabilizing factor in this configuration. Answer-only masking contributes more modestly, and deep-to-shallow injection is stronger than the alternative layer-direction choices.

\begin{table*}[t]
\centering
\caption{Full task-level ablation results on LLaMA3-8B (\%).}
\label{tab:ablation-full}
\resizebox{\textwidth}{!}{%
\begin{tabular}{lccccccccc}
\toprule
Setting & BoolQ & PIQA & SIQA & HellaSwag & WinoGrande & ARC-e & ARC-c & OBQA & Avg. \\
\midrule
A-0 Reproduced LoRA baseline & 66.6 & 89.3 & 81.8 & 95.5 & 87.0 & 91.3 & 81.7 & 84.4 & 84.7 \\
A-1 Full Echo-LoRA & 75.6 & 90.2 & 82.5 & 96.6 & 89.8 & 93.6 & 82.4 & 89.2 & 87.5 \\
A-2 w/o Answer-Only Masking & 75.0 & 90.0 & 81.8 & 96.7 & 89.7 & 93.1 & 83.7 & 87.8 & 87.2 \\
A-3 w/o Stochastic Routing & 72.9 & 81.8 & 82.7 & 92.0 & 34.2 & 93.2 & 84.1 & 88.8 & 78.7 \\
A-4 Deep $\rightarrow$ Deep & 70.6 & 89.6 & 82.8 & 96.5 & 75.4 & 93.0 & 82.6 & 88.2 & 84.8 \\
A-5 Shallow $\rightarrow$ Shallow & 73.4 & 88.5 & 83.0 & 94.5 & 86.8 & 93.7 & 82.3 & 87.0 & 86.2 \\
A-6 \texttt{q\_proj} only & 75.7 & 90.3 & 83.0 & 96.5 & 89.2 & 92.3 & 82.9 & 89.4 & 87.4 \\
A-7 \texttt{v\_proj} only & 75.3 & 90.1 & 82.3 & 96.5 & 88.7 & 93.1 & 82.4 & 88.4 & 87.2 \\
A-8 All attention projections & 75.7 & 90.3 & 82.0 & 96.5 & 89.5 & 93.1 & 83.0 & 90.2 & 87.5 \\
A-9 w/o Answer-Only Masking + all attention projections & 76.0 & 90.7 & 82.1 & 96.4 & 88.1 & 93.1 & 83.6 & 88.0 & 87.3 \\
\bottomrule
\end{tabular}}
\end{table*}

\subsection{Additional Observations}

Echo gains vary across tasks. Datasets such as HellaSwag, WinoGrande, and OBQA require context integration and candidate discrimination, so the injected semantic signal may be more useful there. When the baseline is already strong, the gains are usually smaller. We interpret this pattern as evidence that Echo supplies task-relevant semantic cues beyond the original low-rank path, with the realized benefit depending on task difficulty, backbone capacity, and baseline strength.

Echo also yields larger gains on LoRA than on DoRA. When the underlying PEFT method is strengthened by weight decomposition, the cross-layer signal appears to act more as a robust supplement than as a large shift in the adaptation ceiling. This pattern may help guide future combinations of Echo-style training paths with other PEFT frameworks.

\end{document}